\documentclass[smallcondensed]{svjour3}

\usepackage{amsmath}
\usepackage{amssymb}
\usepackage[hyphens]{url}
\usepackage{latexsym}
\usepackage{tikz}
\title{A fully automatic problem solver with human-style output}
\author{M. Ganesalingam and W.T. Gowers}

\def \imax{\mathrm{imax}}

\def\longline{\vspace{-2mm}\line(1,0){100}\vspace{0.2mm}}
\def\shortline{\line(1,0){60}}
\def\shorterline{\line(1,0){30}}

\begin{document}
\maketitle

\section{Introduction and related work}

\subsection{Overview of the paper}

The main purpose of this paper is to describe a program that solves elementary mathematical problems, mostly but not exclusively in metric space theory, and presents the solutions in a form that is hard to distinguish from solutions that human mathematicians might write. The following two proofs are examples of the program's output.\footnote{The program produces LaTeX output, which we reproduce verbatim here.} The first is a proof that if $f:X\to Y$ is a continuous function and $U$ is an open subset of $Y$, then $f^{-1}(U)$ is an open subset of $X$, and the second is a proof that if $f:X\to Y$ is an injection and $A$ and $B$ are subsets of $X$, then $f(A)\cap f(B)\subset f(A\cap B)$.

\begin{quote}
Let $x$ be an element of $f^{-1}(U)$. Then $f(x)\in U$. Therefore, since $U$ is open, there exists $\eta>0$ such that $u\in U$ whenever $d(f(x),u)<\eta$. We would like to find $\delta>0$ s.t. $y\in f^{-1}(U)$ whenever $d(x,y)<\delta$. But $y\in f^{-1}(U)$ if and only if $f(y)\in U$. We know that $f(y)\in U$ whenever $d(f(x),f(y))<\eta$. Since $f$ is continuous, there exists $\theta>0$ such that $d(f(x),f(y))<\eta$ whenever $d(x,y)<\theta$. Therefore, setting $\delta=\theta$, we are done.
\end{quote}

\begin{quote}
Let $x$ be an element of $f(A)\cap f(B)$. Then $x\in f(A)$ and $x\in f(B)$. That is, there exists $y\in A$ such that $f(y)=x$ and there exists $z\in B$ such that $f(z)=x$. Since $f$ is an injection, $f(y)=x$ and $f(z)=x$, we have that $y=z$. We would like to find $u\in A\cap B$ s.t. $f(u)=x$. But $u\in A\cap B$ if and only if $u\in A$ and $u\in B$. Therefore, setting $u=y$, we are done.
\end{quote}

\noindent The `human-style' output demonstrated by these two examples is the most visible novel feature of the program, since the problems it solves are all well within the capacity of existing programs. However, this novelty reflects a less obvious novelty in the way that the program operates. It would not be able to produce human-style output if it did not mirror very closely the way that human mathematicians think. We envisage that the program will be just the first in a sequence of programs, and we hope that as a result of our detailed attention to human thought processes, future programs in the sequence will be able to solve problems that have not previously been solved by fully automatic provers. 

The structure of this paper is as follows. We begin with a brief discussion of why there has not been more interaction between mathematicians and researchers in automatic theorem proving. Over the following three subsections, we talk about human-oriented theorem proving and explain why we see this tradition as appropriate for what we are trying to do. We then discuss previous work on natural-language output for theorem-proving programs.

In the second section of the paper, we explain how our program works. We start by informally presenting an example of the program in action. This is followed by a description of the `moves' that the program makes. We then present a second example, this time discussing more explicitly why the program does what it does. We end the section by explaining how the program writes up its thoughts to produce the kind of output shown above.

In the third section, we describe an experiment that we conducted in order to test whether the output of the program was hard to distinguish from genuine human mathematical writing. We conclude with a brief discussion of our future plans.

\subsection{Mathematicians' attitudes to automated provers}

We come to this project as mathematicians who are fascinated by the question of how people manage to think of proofs of difficult theorems. Since this question would seem to be fundamental to the whole enterprise of mathematical research, one might think that large numbers of mathematicians would have studied it. However, although mathematicians certainly have thought about their thought processes -- the most famous example being Polya \cite{polya1957solve,polya1954mathematicsi,polya1954mathematicsii} -- far less attention is paid to the topic than one might expect, and papers are routinely written in a style that appears to do its best to conceal how the ideas they contain were discovered.

The result is that the work of analysing and understanding in detail how mathematicians arrive at proofs has to date been mainly carried out by the automatic theorem proving community. Unfortunately, this work has been largely ignored by mathematicians: regrets about this have been expressed in many places in the literature. In a recent article, Bundy discussed this indifference and suggested several reasons for it \cite{bundy2011}. From his reasons one general theme emerges: the mathematical style of current automated provers is very different from the style of human mathematics. This is not just a question of the language such provers use to express problems and their solutions, but also the nature of the solutions themselves, which are often very long and use low-level arguments where a human mathematician would use high-level arguments. Bundy suggests that things are changing (in response to the last point he suggests using `hiproofs', a hierarchical presentation of proofs that allows the reader to look at them at several different levels of granularity) and that it is inevitable that automated provers will in due course become an indispensable tool for mathematicians. We agree with this, and one of our motivations is to do what we can to hasten this development.

What could induce mathematicians to be interested in automated provers? In some cases, the answer is easy: there are mathematics problems that require large searches or enormously complicated calculations, and computers are better at these tasks than humans. But most mathematics is not like that, and for the part that is not, there is a difficulty. Current automated provers are still a long way from being able to answer questions (except of the specific kind just mentioned) that might arise in a typical research project, so the task for the moment is to teach computers to solve much easier questions of a kind that mathematicians can easily do by hand. This task, though it has considerable intrinsic interest, has no immediate payoff for mathematical researchers, so if one wants input from mathematicians who are not specialists in automated theorem proving, then the barrier to entry will have to be very low.

For this reason it would be highly desirable to have an automated prover with the following properties.
\begin{enumerate}
\item (\emph{User-friendly input.}) One can input problems without needing to learn a special-purpose formal language.
\item (\emph{User-friendly output.}) The program will output solutions expressed in the language that mathematicians customarily use.
\item (\emph{Informative solutions.}) The solutions are not just certificates of truth: they are arguments of the kind that a good human mathematician might produce.
\item (\emph{Easy extendability.}) One can increase the power of the program by adding data in the form of facts, problem-solving tips, etc., rather than having to rewrite the entire program.
\end{enumerate}
In short, it would be wonderful to have an automated theorem prover with which one could interact in much the way one interacts with a human mathematician.

This is an ambitious goal that will be difficult to achieve, but if it can be achieved, then it is likely that far more mathematicians will become interested in automatic theorem proving.

\subsection{Human-oriented theorem proving: some background}
\label{sec:history}

Since the early days of automated theorem proving, there have been two competing broad approaches. The first, \emph{human-oriented}, tradition focuses on analysing human methods and replicating them in programs.  The second,  \emph{machine-oriented}, tradition instead relies on the brute strength of computers, using extensive search to solve problems in ways that a human would find difficult to replicate. Given our long-term goal of helping to create a program that mathematicians will want to use, it is natural that our own work should belong firmly in the human-oriented tradition.  However, this tradition is distinctly out of favour at present; for example, if you look at the range of solutions to the standard TPTP test library automated theorem provers \cite{tstp1,tstp2}, you could be forgiven for not realizing that the human-oriented tradition had ever existed.  

Since the human-oriented tradition is less well known these days, we shall briefly describe it here. Our account is not intended to be comprehensive: we focus on the strands that most closely relate to our own work.   More detailed accounts of the various periods in theorem proving may be found in \cite{bledsoe1988survey}, \cite{wos1985overview}, \cite{mackenzie2001}, pp. 464--469 of \cite{harrison2009book}, the introduction to \cite{bundy1983}, \cite{bundy1999survey},  and \cite{cordeschi1996}.

The earliest human-oriented prover was Newell and Simon's 1956 \emph{Logic Theory Machine} \cite{newell1957logic}, which proved theorems from Russell and Whitehead's Principia Mathematica \cite{russell1910}, and is often considered to be the first artificial intelligence program.  Its methods were directly modelled on those of humans: Newell and Simon studied their own behaviour when proving theorems, and abstracted out \emph{heuristics} which they then implemented in their program.  Their goals were wider than those of contemporary theorem proving: they hoped not just to prove theorems but also to `[apply] research on complex information-processing systems ... [to] human learning and problem solving'.  Indeed, they used the successor to the Logic Theory Machine, their \emph{General Problem Solver} \cite{newell1959gps}, as the foundation of a book entitled \emph {Human Problem Solving} \cite{newell1972human}.

In the subsequent decade, much of the work on automated theorem proving, including the General Problem Solver and Gelernter's \emph{Geometry Theorem Proving Machine}, continued in the heuristic, human-oriented tradition, sometimes drawing directly on mathematicians' analyses of their own methods in works such as \cite{polya1957solve,polya1954mathematicsi,polya1954mathematicsii}. This trend was arrested by Robinson's highly influential paper \emph{A Machine-Oriented Logic Based on the Resolution Principle}. Robinson's \emph{resolution} was a single, simple inference rule which turned out to be dramatically more effective than previously existing methods.  Unlike the earlier methods, resolution had no natural human correlate: formulae needed to be converted into a not very readable `normal form' before resolution could be applied, and actual resolution proofs were often difficult or impossible for humans to follow. However, the hugely increased efficiency of resolution meant that these were minor concerns.

The next few years saw an explosion of research into resolution.  Numerous improvements were rapidly found, and it seemed at the time that a suitably advanced version of resolution might solve the problem of automated theorem proving well enough for all practical purposes.  As a result, human-oriented work was more or less abandoned: while a handful of researchers, including  Newell and Simon, remained focused on human cognition, the majority were concerned with problem-solving effectiveness and so devoted their attention to resolution. 

According to \cite{mackenzie2001}, the optimism of those times was somewhat dimmed by the discovery of NP-completeness \cite{cook1971complexity,karp1972reducibility}, and the associated realization that it was unlikely that there was a complete, polynomial time algorithm for theorem proving. Nevertheless, the majority of researchers continued to focus on machine-oriented methods throughout the 70s. The notable exception to this was a group run by Woody Bledsoe at the University of Texas in Austin. Bledsoe's first steps in the direction of human-oriented theorem proving may be found in \cite{bledsoe1971splitting}, which uses top-level heuristic techniques to split a problem into smaller subproblems  which are then solved by resolution.  This paper was soon followed by a 1972 paper on limit theorems \cite{bledsoe1972limit}, written by Bledsoe and his students  Boyer and Henneman. They took an existing system \cite{bledsoe1971splitting} and replaced resolution with a new procedure, which he described as `[bearing] a closer resemblance to the proof techniques of a mathematician than does resolution.'  Among other results, this procedure was able to prove that differentiability implies continuity and that the limit of a product of functions is the product of the individual limits. In the light of the fourth desirable property mentioned above, one feature of the procedure that is worth highlighting is that it `[did] not release its action unless its need is detected'; that is, it did not interfere with those proofs in which was not potentially useful. This is in sharp contrast to resolution, where adding irrelevant axioms can significantly degrade performance. The paper notes that  such heuristics `should be sought for other areas of mathematics'. 

Five years later, Bledsoe wrote a seminal paper on mathematical human-oriented proving, \emph{Non-resolution theorem proving} \cite{bledsoe1977nonresolution}. In it, he described himself as one of a number of researchers who had `made the switch' after finding that resolution and related techniques had encountered enormous difficulty in solving problems that a human would find easy, and that substituting human techniques resulted in a program which `easily succeeded'. 
Bledsoe expressed the opinion that ``purely syntactic" methods such as resolution had reached a plateau, and that further progress would not be made without using more human techniques.  He described the key to progress as somehow managing to use `the knowledge accumulated by humans over the last few thousand years, to help direct the \emph{search} for proofs'.  The paper went on to outline the main concepts used in human-oriented proving at the time. A number of these are very relevant to the work we describe below, including the use of rewrite rules, forward chaining, typing, a reluctance to expand definitions unless necessary, and the use of `natural' or `goal-oriented' systems. 

Bledsoe continued to work on human-oriented proving for the remainder of his life; notable achievements include the proof of a number of results using non-standard analysis \cite{bledsoe1977automatic}, a proof of the intermediate value theorem \cite{bledsoe1977setvariables}, the use of examples to guide proofs \cite{bledsoe1983usingexamples}, and analogical proof construction \cite{bledsoe1995analogy}. Reading through these papers, 
one has no sense that human-oriented proving had reached a plateau: Bledsoe's approach to automating the discovery of proofs of mathematical theorems continued to be fertile throughout his life.  Unfortunately, the `AI winter' meant that funding for mathematical theorem proving dropped sharply soon after Bledsoe `made the switch'.  As a result, many of Bledsoe's students were forced to move to areas with more direct commercial relevance than mathematics, with the result that the Bledsoe tradition in mathematical ATP effectively petered out. Bledsoe's students did, however, carry his human-oriented tradition into other areas of theorem proving.  Of particular note are Robert Boyer and J.\ Strother Moore, who created a family of provers that are used to this day, primarily for software and hardware verification.  

Although many features of the Boyer-Moore provers, NQTHM and ACL2, were driven by the needs of software verification as opposed to mathematics, two features of these provers are significant in a mathematical context. First, Boyer and Moore emphasise the importance of creating a single program that could be used to tackle a range of problems, whereas Bledsoe constructed a sequence of loosely related provers, each intended to tackle different problems.  Second, based on intensive inspection of their own proof techniques, Boyer and Moore introduced an overall architecture, the `waterfall' architecture, which was much more systematic than the architectures introduced by Bledsoe. The Boyer-Moore waterfall consisted of four heuristics, ordered in terms of their relative attractiveness to humans. A Boyer-Moore prover repeatedly applies the least risky heuristic that can be used, terminating when the result is proved or when no heuristic can be applied. 

The Boyer-Moore provers partially inspired work in the next chapter of human-oriented theorem proving, the work done by the group run by Alan Bundy at the University of Edinburgh.  Bundy had been influenced by real, human mathematics from the beginning of his career \cite{bundy1975analysing,bundy1985discovery}, and reports a Bledsoe-like experience of disillusionment with machine oriented methods \cite{bundy1983book}, but his major achievements in human-oriented proving began with a later attempt to reconstruct the behaviour of the Boyer-Moore prover within a more systematic framework \cite{bundy1988proofplans}. This led him to formulate the notion of a \emph{proof plan}, a computational representation of a high-level heuristic used by human mathematicians.  Because one application of a proof plan corresponded to many smaller reasoning steps, a theorem prover that used proof plans could operate with a dramatically reduced search space. This often led to a proof discovery process that was very similar to that of a human mathematician. There is an extensive literature on proof plans, which we will not discuss here, since proof plans are not explicitly used in the work described below. A description and evaluation of this work may be found in \cite{bundy2000critique}. 


Meanwhile, on the machine side there was a notable triumph: a result with a strong claim to be the first solution by an automated theorem prover of an unsolved problem in mainstream mathematics. The problem was the Robbins conjecture, that an algebra that obeyed a certain axiom must be a Boolean algebra, which had been open for six decades when William McCune proved it in 1996 using an automated theorem prover called EQP (for `equational prover') and eight days of computer time \cite{mccune1997solutionof}.

This might appear to indicate that the two aims of producing more powerful provers and of producing provers that will be easy to use by human mathematicians are in conflict. We do not believe that this is the case, for reasons that we shall now explain.

\subsection{Doing more with less}

As we have already said, if one wants to write a program with which mathematicians can interact easily, it is natural to be drawn to the human-oriented tradition. However, as the history of that tradition makes clear, that is by no means the only motivation. There is also a conviction amongst proponents of human-oriented methods that in order to build provers that can solve certain kinds of problems -- roughly speaking, the problems that human mathematicians typically deal with, apart from the ones that need a lot of search or heavy calculation -- it is \emph{necessary} to come to grips with human problem-solving techniques. For example, Bundy says the following in his influential \emph{Science of Reasoning}.

\begin{quote}
Although our science of reasoning might find application in the building of high performance automatic theorem provers, the two activities are not co-extensive. They differ both in their motivation and their methodology.

I take the conventional motivation of automatic theorem proving to be the building of theorem provers which are empirically successful, without any necessity to understand why. The methodology is implied by this motivation. The theorem prover is applied to a random selection of theorems. Unsuccessful search spaces are studied in a shallow way and crude heuristics are added which will prune losing branches and prefer winning ones. This process is repeated until the law of diminishing returns makes further repetitions not worth pursuing. The result is fast progress in the short term, but eventual deadlock as different proofs pull the heuristics in different directions. This description is something of a caricature. No ATP researchers embody it in its pure form, but aspects of it can be found in the motivation and methodology of all of us, to a greater or lesser extent.

Automatic theorem provers based on proof plans make slower initial progress. Initial proof plans have poor generality, and so few theorems can be proved. The motivation of understanding proofs militates against crude, general heuristics with low prescriptiveness and no expectancy. The `accidental' proof of a theorem is interpreted as a fault caused by low prescriptiveness, rather then a lucky break. However, there is no eventual deadlock to block the indefinite improvement of the theorem prover's performance. If two or more proof plans fit a theorem then either they represent legitimate alternatives both of which deserve attempting or they point to a lack of prescriptiveness in the preconditions which further proof analysis should correct.

[...]

Thus, we expect a science of reasoning will help us build better automatic theorem proving programs in the long term, although probably not in the short term. 
\end{quote}

\noindent Although Bundy's `science of reasoning' was (at least initially) tightly focused on the notion of a proof plan, we believe that many of his comments apply more generally to the advantages of human oriented proving over machine oriented proving, and especially his conclusion: that in the long term, paying close attention to human methods will pay dividends.

There are two main reasons for this. A simple one is that a significant problem with automated theorem provers -- some would call it \emph{the} main problem -- is the danger of combinatorial explosion. It is in the nature of solving a complex mathematics problem that one throws up other problems that need solving, which in turn throw up further problems, and so on. If a generous amount of search is permitted, then this recursive nature of problem solving naturally leads to the search being iterated, and thus to a combinatorial explosion. Somehow, in ways that we do not fully understand, humans manage to avoid this difficulty by keeping search strictly under control. Therefore, it seems highly plausible that it will not be possible to design provers that can produce complicated hierarchical proofs without a deep understanding of how humans manage to do so.

The second reason is more subtle, and best illustrated by means of an example. Consider the following problem.

\begin{problem} Let $G$ be a group and let $S$ be a subset of $G$ with the property that if $x\in S$ and $y\in S$, then $xy^{-1}\in S$. Prove that $S$ is closed under taking inverses. 
\end{problem}

\noindent This was used as a test problem for a system devised by Reiter in the 1970s \cite{reiter1976models}. His system was designed to use models to prune search trees: if a statement is false in a simple model, then there is no need to waste time investigating whether it can be proved. It is clear that this reflects an extremely important aspect of human problem solving: before we invest time in proving a statement, we like to feel that that statement has at least some plausibility. Therefore, it is natural to want to incorporate some kind of checking-against-models facility into a theorem proving program. 

Here, translated for human consumption, is what Reiter's system does. It begins by taking an element $b\in S$ (if $S$ is empty then the problem is trivial) and sets itself the goal of proving that $b^{-1}\in S$. The one property it has available for use is that $S$ is closed under the operation $(x,y)\mapsto xy^{-1}$, so it uses backwards reasoning to replace the goal by that of finding $x,y\in S$ such that $xy^{-1}=b^{-1}$. 

At this point, it searches for suitable candidates for $x$ and $y$. It has three statements that these two variables need to satisfy: $x\in S$, $y\in S$ and $xy^{-1}=b^{-1}$. It searches for a solution in a somewhat mechanical way, ordering the conditions that need to be satisfied and then attempting to satisfy them one by one. It begins with the order in which the conditions are presented, so it tries to satisfy the first condition. Since all it knows so far is that $b\in S$, it takes $x=b$. The first statement is now satisfied, so it passes to the second and takes $y=b$ for similar reasons. It is now left needing to prove that $bb^{-1}=b^{-1}$. 

Instead of wasting time looking for a proof of this false (in general) statement, it checks it against a model such as the Klein four-group, rapidly finding a counterexample. So at this point it backtracks, reorders the statements that need to be satisfied, and ends up attempting to satisfy what used to be the third statement, $xy^{-1}=b^{-1}$, without, for the time being, worrying about whether $x\in S$ and $y\in S$. It finds the simplest solution to this, namely $x=e$ and $y=b$. Now it is left needing to prove that $e\in S$, which is straightforward: it needs $u,v\in S$ such that $uv^{-1}=e$ and the first thing it tries, namely $u=v=b$, works.

Thus, Reiter's system successfully solves the problem, and even does so quite efficiently. However, there are at least two ways in which what it does is strikingly unlike what a human mathematician would do. The first is its mechanical approach to satisfying the conditions on $x$ and $y$. In general, finding an object that satisfies two or more properties is a tricky problem, since it is difficult to focus on more than one condition at a time. Humans will often begin by assessing which is the hardest condition to satisfy, since in practice it often happens that the most natural object that satisfies the hardest condition either already satisfies the easier conditions or can be easily modified to do so. By contrast, Reiter's program starts by attempting to satisfy the conditions in the order in which they are written down, which leads it to make a wrong guess that no human mathematician would be likely to make. 

If we are just interested in the performance of one program, this is not much of a problem, since there are only three conditions, so the amount of backtracking it has to do is very slight. However, if we are content with this in the short term, then we miss out on search-reducing techniques that may be useful in the longer term.

The second noticeably non-human move made by the computer is to test the statement $bb^{-1}=b^{-1}$ against a model (in Bledsoe's account of this work, he suggests using the Klein 4-group). To a typical human mathematician, this statement is just \emph{obviously} false in general. Of course, it is not much help to a theorem proving program to be told that a statement is `just obvious', so let us look more closely at why it is obvious. The answer is that we see very easily that, by the cancellation law, if $bb^{-1}=b^{-1}$ then $b=e$. And even at that point we say to ourselves something like, ``Not all groups are trivial," rather than ``The cyclic group of order 2 is a counterexample." Thus, if we are content with Reiter's method, then we miss out on two further search-reducing strategies: to simplify statements before deciding whether they are likely to be true, and to make the working assumption that a sufficiently simple general statement that is not obviously true is almost certainly false. 

That is not all that can be extracted from a close look at how a human would tackle this problem. Reiter's system always reasons backwards, but a human might well look at the hypothesis that $S$ is closed under the operation $(x,y)\mapsto xy^{-1}$ and reason that since the only element we know to be in $S$ is $b$, there is no hope of solving the problem without making the deduction that $bb^{-1}$ is an element of $S$, and therefore that $S$ contains the identity $e$. And once one has made that deduction, finding $x$ and $y$ in $S$ such that $xy^{-1}\in S$ is significantly easier than it was before. So we have another potentially useful technique: if a statement appears to be essential and can be applied in only one way, then there is no harm in applying it, even if you cannot see what good it will do. 

These remarks are not intended as a criticism of Reiter's program, which was successful at what it was trying to do. But they illustrate the general point that the closer the attention one pays to how humans solve problems, the more useful information one can extract. This gives us another reason for trying to make our programs as human as possible. Indeed, we would describe our work as belonging not just to the human-oriented tradition, but as belonging to the extreme human end of the machine-human spectrum. In practice, this means deliberately not allowing ourselves to exploit the speed of computers, for example by letting them carry out large searches or perform very complicated calculations. We hope that by submitting to this restriction, we will force ourselves to develop a number of useful and important techniques while the problems we are tackling are still relatively simple.

\subsection{Two kinds of proof}

The arguments of the previous section are intended to show that there are potential advantages to taking a strict human-oriented approach to automated theorem proving. However, there is no denying that there are also advantages to using raw computer power. And the fact that it is the machine-oriented approach that has solved a decades-old unsolved research problem suggests that its advantages outweigh those of the human-oriented approach. Perhaps the situation with theorem proving is like the situation with chess, where the brute force of the computer beats the advanced pattern-recognition skills of the human grandmaster.

Before we jump to that conclusion, however, we should consider another possibility: that machine-oriented methods are better for finding certain kinds of proofs, and human-oriented methods are better for finding other kinds. The evidence suggests that this is the case. As Kerber puts it \cite{kerber1998}, when talking about machine-oriented programs, ``The strength of these systems is truly remarkable and even open mathematical problems, like recently the question whether Robbins algebras are Boolean algebras, can be solved with the assistance of an automated theorem prover. On the other hand, observing the blind search behaviour of such a system as it fails to solve a problem that seems trivial to us as humans can be disappointing."

This phenomenon is not too surprising. Some proofs seem to consist of a succession of somewhat arbitrary and unpredictable steps, while others can be discovered by means of what mathematicians would describe as `key ideas'. The proof of the Robbins conjecture comes into the first category: one can think of it as a combinatorial object that belongs to a vast space -- the space of sequences of well-formed statements that satisfy certain simple rules -- inside which we are looking for a sequence that ends in a certain way. There are no obvious measures of progress that tell us that some initial segments of sequences are `obviously right' or at least `getting warmer' (though it is conceivable that somebody might one day find a more conceptual argument), and in the absence of such clues there is not much for it but to undertake a huge search. It is therefore only to be expected that machine-oriented methods have outperformed human-oriented methods for that problem and for others of a similar kind.

However, for the majority of proofs that mathematicians find, there is some kind of `story' to tell of the ideas that give rise to the proof. Typically, such a story will be a high-level overview of the main difficulty and how it is overcome, where `overcome' means that the problem is reduced to one or more problems where that difficulty no longer occurs. Often this reduction is achieved by means of a well-chosen intermediate statement that turns out to follow from the initial assumptions and imply the conclusion. The intermediate statement itself is typically found not by means of a brute-force search but by a process of approximation: one might make a guess, find that it is unhelpful, understand why it is unhelpful, and use that understanding to guide the search for a better intermediate statement. These characteristically human techniques enable mathematicians to penetrate deep into `proof space', but the set of proofs that can be discovered in this way forms a tiny fraction of that space. It seems almost a truism that human methods will be useful for programs that want to find these special proofs that human mathematicians are so mysteriously good at finding.

\subsection{Systems with natural-language output}

So far, we have discussed the desirability of a program that will reason in as similar a way as possible to the way that humans reason. Another desirable property that we mentioned earlier is that the input and output of the program should be in the language that mathematicians already use. Several systems have been developed that use natural language to a greater or lesser extent. 

An early example with some similarity to ours is that of Felty and Miller from 1987 \cite{feltymiller1987}. They start with a proof tree and convert it into a more readable form. Their system can also make significant changes to how a proof is presented. The following is an example of output from their system: it is a proof that there are infinitely many primes. The function $f$ mentioned in the proof can be taken to be the function defined by the formula $f(n)=n!+1$: then the beginning of the proof is asserting some properties of this function that are subsequently used in the proof (so any other function with those properties would do just as well). 

\begin{quote}
Assume $\forall x (f(x)>x)\wedge\forall x\forall y(\mathrm{div}(x,f(y))\supset(x>y))\wedge\forall x(\neg\mathrm{prime}(x)\supset\exists y(\mathrm{prime}(y)\wedge\mathrm{div}(y,x))$. We have two cases. Case 1: Assume $\neg\mathrm{prime}(f(a))$. By modus ponens, we have $\exists y(\mathrm{prime}(y)\wedge\mathrm{div}(y,f(a)))$. Choose $b$ such that $\mathrm{prime}(b)\wedge\mathrm{div}(b,f(a))$. By modus ponens, we have $(b>a)$. Hence, $(b>a)\wedge\mathrm{prime}(b)$. Thus, $\exists x((x>a)\wedge\mathrm{prime}(x))$. Case 2: Assume $\mathrm{prime}(f(a))$. Hence, $(f(a)>a)\wedge\mathrm{prime}(f(a))$. Thus, $\exists x((x>a)\wedge\mathrm{prime}(x))$. Thus, in either case, we have $\exists x((x>a)\wedge\mathrm{prime}(x))$. Since $a$ was arbitrary, we have $\forall n(\exists x((x>n)\wedge\mathrm{prime}(x)))$. 
\end{quote}

They describe their mechanism for converting the original tree-structured deductions into readable natural-language text as very simple. It is clear that with some small changes they could have improved the readability. For example, they could have replaced $\mathrm{prime}(x)$ by `$x$ is prime', $\mathrm{div}(x,y)$ by $x|y$ and the symbols for connectives by English words. However, the result would still have had some slightly odd characteristics -- for instance, no human mathematician would bother to write `by modus ponens' -- that would have betrayed its mechanical origins.


Another program that produced readable text was written by Holland-Minkley, Barzilay and Constable in 1999 \cite{hollandminkley1999verbalization}. Their aim was to create natural-language output from the Nuprl system. This is an interactive system based on tactics -- that is, high-level inference steps -- which is designed to mimic human reasoning. The output from the Nuprl system is not at all easy for the untrained mathematician to read. However, they could convert it into language that was considerably closer to what a human mathematician might write, as the following sample demonstrates. 
\bigskip

\begin{quote}
Theorem: For integers $a$ and $b$ and natural number $c$, 
$(a --b) --c = a --(b + c)$. 
\smallskip

Consider that $a$ and $b$ are integers and $c$ is a natural number. Now, the original expression can be transformed to $\imax(\imax(a - b; 0) - c; 0) = \imax(a - (b + c); 0)$. From the add com lemma, we conclude $\imax( -c+\imax(a + -b; 0); 0) =\imax(a + -b + -c; 0)$. From the imax assoc lemma, the goal becomes $\imax(\imax((a + -b) + -c; 0 + -c); 0) = \imax(a + -b + -c; 0)$. There are 2 possible cases. The case $0 + -c \leq 0$ is trivial. Consider $0 < 0 + -c$. Now, the original expression can be transformed to $\imax((a + -b) + -c; 0 + -c) = \imax(a + -b + -c; 0)$. Equivalently, the original expression can be rewritten as $\imax((a + -b) +-c) = \imax(a +-b +-c; 0)$. This proves the theorem. 
\end{quote}
\bigskip

In many ways this looks like the kind of continuous prose that a mathematician would write, though as with Felty and Miller's system there are a number of telltale signs of the mechanical origins of the text. For instance, the first sentence is not quite grammatical: a human would write, `Let $a$ and $b$ be integers and let $c$ be a natural number.' There is also the trivial point that mathematicians would write `max' rather than `imax' (trivial because it would be very easy to change this). There is also a repetitive quality to the prose that gives it an automatically generated feel: for instance, two sentences open with `Now, the original expression can be transformed to'. 



Another system that is often promoted for the readability of its output is MIZAR. Although this is not an automatic theorem prover, we show an example of its output for the sake of comparison. The following is a formalized version in MIZAR of Euclid's proof that there are infinitely many primes. [From \url{http://www4.in.tum.de/~wenzelm/papers/romantic.pdf} .]
\medskip

\begin{tt}
reserve n,p for Nat; 

theorem Euclid: ex p st p is prime \& p > n 

proof 

set k = n! + 1; 

n! > 0 by NEWTON:23; 

then n! >= 0 + 1 by NAT${}_{}$1:38; 

then k >= 1 + 1 by REAL${}_{}$1:55; 

then consider p such that 

A1: p is prime \& p divides k by INT${}_{}$2:48; 

A2: p <> 0 \& p > 1 by A1,INT${}_{}$2:def 5; 

take p; 

thus p is prime by A1; 

assume p <= n; 

then p divides n! by A2,NAT${}_{}$LAT:16; 

then p divides 1 by A1,NAT${}_{}$1:57; 

hence contradiction by A2,NAT${}_{}$1:54; 

end; 

theorem {p: p is prime} is infinite 

from Unbounded(Euclid); 
\end{tt}
\bigskip

\noindent Although this is quite clearly written in a formal language, it is natural enough that with a bit of effort, one can follow the steps of the argument, at least if one knows the argument already. And if readability were the only concern, then once again there would be simple ways of improving the output. For example, one could remove all the `by X' justifications at the ends of lines. Of course, readability is \emph{not} the only concern when a proof is written in MIZAR. More generally, there are major differences between our aims and the aims of those who formalize mathematical proofs. Our aim is to write fully automatic programs with output that meets the typical standards of soundness of human mathematicians, whereas the proof-verification community aims to use a great deal of human interaction to produce completely formalized proofs. 
\medskip

As we have already mentioned, we would ultimately like to contribute to the creation of a program that mathematicians can use with little or no effort. It therefore makes sense to set very high standards for the readability of our output. Another reason for doing this is that we would like to be confident that our programs really are thinking in a human way. A natural test of this is that it should be able to produce output that is similar to what a human produces. 

However, passing that test is not by itself sufficient evidence, since in principle one might be able to create a program that discovers human-style proofs with the help of a highly mechanical and non-human discovery process. Such a program could then hide its thought processes and produce nice readable write-ups. We therefore want to add an extra condition: we would like our programs to create human-style write-ups \emph{while being faithful to its thought processes}. By this we mean that the steps that the program takes in order to discover the proofs it discovers should be directly translated into human language and should be presented in the order in which they are made. For example, if a program discovers a short argument after a long search, then the long search should be presented and not just the short argument. 

There is an obvious objection to this, which is that not even human mathematicians do it. For example, often they will take all sorts of wrong turns when searching for proofs, and these will not be recorded in the write-ups. And often they will present, as if out of nowhere, an object that just happens to make a proof work, hiding from view the process by which they thought of it. 

This is an important point, but the proofs we will discuss in this paper are not sufficiently difficult to find for the issue to arise in a significant way. For the proofs the program finds, it is possible to produce output that is faithful to the program's thought processes and also quite close to what a human would write as the final product. When we come to tackle more complex problems, we plan to have two styles of output. One will be a proof as a human might write it, and the other will be more like an account, again as a human might write it, of the proof discovery process. We believe that both styles are potentially useful to mathematicians.

It is one thing to claim that a program produces output that is similar to what a human might write; it is another to provide evidence to back up that claim. Later in the paper, we shall describe an experiment that we carried out for this purpose.

\subsection{Soundness}

A further motivation for producing human-style write-up is that it gives us a mechanism for checking soundness: if the program produces proofs that a human mathematician finds acceptable, then it is doing what we ask of human mathematicians.

This attitude to soundness is in sharp contrast with most research in automated theorem proving, especially of the interactive variety. There what is sought is a \emph{guarantee} of soundness, which is obtained by carefully building up a corpus of statements from a small set of axioms, using only basic rules of logic and statements that have already been established. There are very good reasons for this: for example, such systems can handle extremely complicated proofs that are very difficult for humans to check by hand, a notable recent example being Gonthier's machine-checked proof of Thompson's odd-order theorem \cite{gonthier2013machine}. They can also check the soundness of computer hardware, where an absolute guarantee is essential.

However, for a program that produces proofs of relatively easy theorems, considerations of this kind do not apply. Therefore, we take the attitude to soundness that a typical human mathematician takes: we want it to be clear to a human mathematician that the steps the program takes are sound, and we want the results that the program assumes to be ones that are well established and in some sense `prior' to the results being proved. Although this does not provide a formal guarantee of soundness, there are no examples of well established mathematical results of the level of simplicity we are dealing with that have been found to be fundamentally wrong, so for all practical purposes it is a sufficient assurance of soundness.

\subsection{The other criteria: user-friendly input and easy extendability}

The remaining two desirable criteria that a mathematician-friendly program should have were that it should be possible to input problems without having to learn a formal language, and that it should be possible to improve the performance of the program by adding data. 

It is much harder to write a program that accepts natural-language input than it is to write a program that produces natural-language output, since for any given problem there will be many ways of writing it, and the program has to be able to handle all of them -- or at least, enough of them to make the program not too restrictive. For the time being, our input takes a form that is not written in natural language, but is fairly straightforward to write. (For example, to request that the program show that the pre-image of an open set $U$ under a continuous function $f$ is open, we supply hypotheses {\tt continuous(f)} and {\tt open(U)} and a target statement {\tt open(preimage(f,U))}. A more complicated example is {\tt forall x epsilon.(in(x,B) => exists y.(in(y,A) \& lessthan(d(x,y), epsilon)))}.) However, Barnet-Lamb and Ganesalingam have recently written a program that is capable of processing many different statements. Roughly speaking, any mathematical statement that is written at the level of formality that is typical of the more formal parts of mathematical textbooks has a good chance of being understood. So it may be that future versions of the program will be able to accept natural-language input as well.

As for the possibility of making the program more powerful by adding more data to it, we are optimistic about this, but it would be premature to make any claims in this direction. In general, adding mathematical information to a theorem-proving program can be problematic, because it gives the program more choice. If the program makes significant use of search, then more choice can make the searches bigger, and in practice this has meant that some programs have performed \emph{worse} when they have more information to go on. The program has not had this problem, but its library is so small that we cannot at this stage claim that it never will. The reason we are optimistic is that our methodology, if we apply it sufficiently rigorously, should ensure that the problem does not arise. If a program threatens to get bogged down searching through all sorts of irrelevant results, then that will be a sign that we have not understood how human mathematicians select relevant results from the large amount that they know. Understanding that may be very challenging, but if we cannot fully understand it, then we will try to reach a partial understanding and make use of that. In other words, we will try to find sufficient conditions for a use of the program's data to be human-like, and allow uses that satisfy those conditions. If the program only ever uses its data in a human way, then adding data should not degrade its performance, since it does not degrade the performance of human mathematicians.

\section{The program}

\subsection{Solving routine problems in a fully human way}

The `extreme human' approach we are trying to adopt can be summarized as follows: \emph{we do not allow our programs to do anything that a good human mathematician wouldn't do}. A serious difficulty in trying to design a program that satisfies this constraint is that while humans do their very best to avoid search and backtracking, they undoubtedly do at least \emph{some} search and backtracking when they are trying to solve difficult problems. What is the distinction between the kind of search that humans do and the kind that humans would never do? Indeed, \emph{is} there a distinction, or is it just a matter of degree?

These are difficult questions, and rather than try to answer them immediately, it seems more sensible to try to isolate them by concentrating first on other difficulties and facing up to this one only when we are forced to do so. Accordingly, we define a \emph{routine} problem to be one that a good human mathematician will typically solve easily without backtracking. If a program is to satisfy the main constraint, then it too will have to solve routine problems without backtracking, so we can simply ban it. Although this is a significant scaling down of ambition, it also makes the project far more realistic in the short term. Furthermore, the class of routine problems is large and diverse enough that the challenges it raises are still very interesting.

\subsection{Example: closed subsets of complete metric spaces}

Before we describe our program, let us look at a routine problem and
examine how a human mathematician would typically solve it. The
problem is to show that a closed subset of a complete metric space is
complete. We can state it more formally as follows.

\begin{problem}
Let $X$ be a complete metric space and let $A$ be a closed subset of
$X$. Prove that $A$ is complete.
\end{problem}

\noindent The proof discovery process would usually be something like this.

\begin{enumerate}
\item \emph{[Clarify what needs to be proved.]} We must show that
every Cauchy sequence in $A$ converges in $A$.
\item \emph{[We must show something about \emph{every} Cauchy
sequence, so pick an arbitrary one.]} Let $(a_n)$ be a Cauchy sequence
in $A$.
\item \emph{[Clarify what now needs to be proved.]} We are trying to
show that $(a_n)$ converges in $A$.
\item \emph{[See what we can say about the sequence $(a_n)$.]} The
sequence $(a_n)$ is a Cauchy sequence in the space $X$, and $X$ is
complete; therefore $(a_n)$ converges in $X$.
\item \emph{[Give a name to the object that we have just implicitly
been presented with.]} Let $x$ be the limit of the sequence $(a_n)$.
\item \emph{[See what we can say about $x$.]} But $A$ is closed under
taking limits, so $x\in A$.
\item \emph{[Recognise that the problem is solved.]} Thus, $(a_n)$
converges in $A$, as we wanted.
\end{enumerate}

Our program is designed to imitate these typical human moves as
closely as possible. The result is that what it does can be
straightforwardly translated into an account of its thought processes
that could pass for an account of human thought processes such as the
ones laid out above (without the accompanying commentary).

The architecture in which moves are applied is very similar to that of a
LCF-style interactive theorem prover \cite{gordon1979lcf}.  In simple
cases, the program state consists of
a list of statements that can be assumed, which we call
\emph{hypotheses}, and a list of statements to be proved, which we
call \emph{targets}. Each statement is 
basically a formula of the (many-sorted) first-order predicate
calculus, but it can also carry annotations that record
information that has been accumulated by the program; for example, a
statement may be tagged to indicate that it has already being used
during the course of the proof.  Although information of this kind is
logically unnecessary, it is indispensable in human reasoning and
therefore plays an important role in our automated theorem
prover.

The representation diverges from that of a LCF-style prover in more
complicated cases, most notably when a conjunctive target needs to be
deduced from the hypotheses. In this case a LCF-style prover would
use a tactic such as replacing the original goal by two new goals, each
of which lists the hypotheses separately. By contrast, our system 
represents and displays the hypotheses only once.  This is not done
for reasons of efficiency, but rather to mimic the natural human
representation.  A human would think in terms of one ambient
primary collection of `facts that are known' (the hypotheses) and a
number of targets that need to be deduced from these; the program
reflects this. In some cases, reasoning will introduce a fact that may
be used in the proof of only one of the targets (and the program represents
this appropriately), but this is very much the marked (i.e.
non-default) case and should not affect the representation of the
majority of `universally available' hypotheses.

An individual \emph{move} is an operation that transforms a specific
problem state into another state in a sound fashion;  thus individual
moves correspond to application of tactics to a specific LCF-style
prover state. However, we insist that our program's moves faithfully model
cognitive processes of human mathematicians. Thus, many sound
transformations of the problem state are not acceptable as moves,
because they correspond to operations that humans would never perform.

The closest correlate we have to a tactic (as opposed to the
application of a tactic to a specific state) is a  \emph{generator}
for a \emph{move type}.  An example of a \emph{move type} would be
`forwards reasoning'.  Each move type has an associated
\emph{generator} which, given a state, returns a list of moves from
that state.  So for example the generator for the `forwards reasoning'
move type accepts a state and returns a list of all the `forwards
reasoning' moves that can be made from that state.  Note that, unlike
tactics, generators may return a whole list of moves.

Because the prover is fully automated, the list of move types is
fixed. Move types are ranked in order of attractiveness or
\emph{priority}, and the basic operation of the program consists of
repeatedly choosing the most attractive move type that can be applied,
generating the moves of that type, and applying the most attractive one. In 
practice with the problems we are considering, it is rare for the most attractive 
move type to generate more than one move, except in cases where the 
moves generated are related by some simple symmetry.

Note that this architecture
closely resembles the Boyer-Moore `waterfall' architecture
\cite{boyermooreacl} discussed in \S\ref{sec:history}. 
The waterfall consisted of four heuristics, ranked in order of their
relative attractiveness to humans. A Boyer-Moore prover repeatedly
applies the most attractive heuristic that can be used, terminating
when the result is proved or when no heuristic can be applied. This directly corresponds to the way in which our prover repeatedly chooses the most attractive move type possible, terminating when the result is proved or when no no moves can be made. 

We shall now describe what our program does when it tackles the
example problem given above, and then explain in more detail what the
move types are, and what the order of priority is. Like Boyer and
Moore, we chose the move types and their priority by examining our own
reactions to many different problems.

\medskip

\noindent The initial problem state for this problem is as follows.
\bigskip

H1. $X$ is a complete space

H2. $A$ is closed in $X$

\longline

T1. $A$ is complete 
\bigskip

\noindent The first thing the program does is expand the definition of ``$A$ is complete". 
\bigskip

H1. $X$ is a complete space

H2. $A$ is closed in $X$

\longline

T2. $\forall (a_n)$ $(a_n)$ is Cauchy $\wedge\ (a_n)$ is a sequence in $A\Rightarrow(a_n)$ converges in $A$
\bigskip

\noindent Next, it does a move that corresponds closely to the human move of picking an arbitrary Cauchy sequence: it gets rid of the universal quantifier and places the premises of the resulting conditional statement above the line, leaving the conclusion below the line. A small technical point is that we insist that it combines these two operations into a single move, because we do not allow ``bare" conditionals: that is, statements of the form $P(x)\Rightarrow Q(x)$ that are not universally quantified. (One reason for doing this is that such statements almost always seem very strange and unnatural to humans. However, we have also found situations where it helps the program to avoid doing genuinely bad moves.)
\bigskip

H1. $X$ is a complete space

H2. $A$ is closed in $X$

H3. $(a_n)$ is Cauchy 

H4. $(a_n)$ is a sequence in $A$

\longline

T3. $(a_n)$ converges in $A$
\bigskip

\noindent Next comes an operation that fits under the general heading of \emph{forwards reasoning}: from the assumptions that $(a_n)$ is Cauchy and $X$ is complete, it follows that $(a_n)$ converges. Of course, this deduction can be broken up: one would expand the statement ``$X$ is complete" to say that every Cauchy sequence in $X$ converges, then substitute $(a_n)$ for the universally quantified sequence, and finally use modus ponens to deduce that $(a_n)$ converges. However, a human mathematician would do all this in one step, and therefore so does our program. Of course, to do it the program must have \emph{access} to the expansion of ``$X$ is complete", which indeed it does: it has a library of definitions and basic facts that it can make use of whenever it wants. This models the collection of definitions and facts that a human mathematician would store in his or her long-term memory. 

After the step, the problem state is this (except that there will be tags on the statements H1 and H3 to indicate that they have been used -- to keep the problem states easy to read we are not displaying the tags in this discussion).
\bigskip

H1. $X$ is a complete space

H2. $A$ is closed in $X$

H3. $(a_n)$ is Cauchy 

H4. $(a_n)$ is a sequence in $A$

H5. $(a_n)$ converges

\longline

T3. $(a_n)$ converges in $A$
\bigskip

\noindent At this point, it will be clear to a human that the hypothesis H1 has been `used up'. That is, it has played its role and will almost certainly not be required again. It is not easy to work out precise necessary and sufficient conditions for when humans judge a statement to be used up in this sense, but we have identified some conditions that appear to be sufficient. The hypothesis H1 satisfies those conditions and is therefore deleted. 

A human mathematician would also be confident that hypothesis H3 has been used up. Our program does not delete H3 because it does not know that the Cauchy assumption will not be expanded later. This we regard as an imperfection of the program that needs at some point to be corrected. (One possible justification for deleting H3 is that it is implied by H5. Whether this is the `right' reason remains to be worked out.)

Deleting statements turns out to have no effect on the performance of this program, so there might seem to be no point in worrying about it. However, we remain of the view that we should try to model \emph{all} aspects of human problem solving. A potential benefit of thinking about the discarding of hypotheses is that it is a special case of a more general issue that undoubtedly \emph{will} be important for performance when the problems get harder: dealing with irrelevant statements. Often with complicated problems one is presented with more information than one needs, and it is important to decide which out of all the statements available are likely to be useful.

\bigskip

H2. $A$ is closed in $X$

H3. $(a_n)$ is Cauchy 

H4. $(a_n)$ is a sequence in $A$

H5. $(a_n)$ converges

\longline

T3. $(a_n)$ converges in $A$
\bigskip

\noindent The next move applied by the program is to expand hypothesis H5. The reason the program likes this move is that the expansion of the hypothesis begins with an existential quantifier, so expanding it gives us a new object to work with. The program automatically removes the existential quantifier, thereby implicitly picking a limit for the sequence $(a_n)$ -- again, this surreptitious removal of quantifiers mirrors the way human mathematicians think and write. After the expansion and quantifier removal, we arrive at the following problem state.
\bigskip

H2. $A$ is closed in $X$

H3. $(a_n)$ is Cauchy 

H4. $(a_n)$ is a sequence in $A$

H6. $(a_n)\to a$ 

\longline

T3. $(a_n)$ converges in $A$
\bigskip

\noindent Now the program does another kind of forwards reasoning, where a result from the library is used. The library result states that a closed set contains its limit points. If we apply this result to the set $A$, the sequence $(a_n)$ and the limit $a$, then the hypotheses H2, H4 and H5 give us precisely the premises we need, which allows us to conclude that $a\in A$.
\bigskip

H2. $A$ is closed in $X$

H3. $(a_n)$ is Cauchy 

H4. $(a_n)$ is a sequence in $A$

H6. $(a_n)\to a$ 

H7. $a\in A$

\longline

T3. $(a_n)$ converges in $A$
\bigskip

\noindent At this point our deletion rules allow the program to delete H2. We also see why it is important to be cautious about deleting statements: the hypotheses H4 and H6 have been used, but they are going to be used again. Fortunately, our deletion rules do not cause either of them to be deleted.
\bigskip

H3. $(a_n)$ is Cauchy 

H4. $(a_n)$ is a sequence in $A$

H6. $(a_n)\to a$ 

H7. $a\in A$

\longline

T3. $(a_n)$ converges in $A$
\bigskip

\noindent The final step for the program is to look at the expansion of the target, which~is
\medskip

$(a_n)$ is a sequence in $A\ \wedge\ \exists z\ z\in A\ \wedge\ (a_n)\to z$
\medskip

\noindent and observe that if it sets $z$ equal to $a$, then all three resulting statements occur above the line as hypotheses. Because of this, it declares the problem solved.

\subsection{Some terminology}

With the possible exception of our rules for deleting statements, which are not essential to the program, all the move types we use in the program are fairly standard in the field, or are combinations and minor variations of standard move types. 

Before we say what they are, it will be useful to have some terminology for the different kinds of statements that can occur. We call a statement \emph{atomic} if it is of the form $P(x_1,\dots,x_k)$, where $P$ is a predicate and $x_1,\dots,x_k$ are terms. We do not insist that it is actually written in this form, so for example the statement 
\medskip

$d(x,y)<\epsilon$ 
\medskip

\noindent is an atomic statement, which we could if we wanted to write in the form 
\medskip

is\_less\_than$(d(x,y),\epsilon)$. 
\medskip

\noindent When a statement is built out of atomic statements using connectives and quantifiers, we classify it according to the operation that appears at the top of its parse tree. For example, the statement 
\medskip

$\exists x\ x\in A\ \wedge\ d(x,y)<\epsilon$
\medskip

\noindent is an \emph{existential} statement, whereas the statement
\medskip

$x\in A\ \wedge\ d(x,y)<\epsilon$
\medskip

\noindent is \emph{conjunctive}. It is often useful to look more than one level down the parse tree. For example, we would call the statement
\medskip

$\forall x\ x\in A\Rightarrow x\in B$
\medskip

\noindent a \emph{universal conditional} statement. (Recall that we do not allow `bare' conditional statements, so this is a particularly important category.) Similarly, the existential statement above can be further classified as an \emph{existential conjunctive} statement. 

Finally, many atomic statements can be expanded into statements that are no longer atomic. For example, the statement $A\subset B$ expands to the universal conditional statement above. It is often useful to know what a statement will become after it is expanded: to specify this we use the prefix `pre-'. Thus, the statement $A\subset B$ is pre-universal conditional. We call an expansion \emph{elementary} if it does not introduce a quantifier. For example, the expansion of $A\subset B$ is not elementary, whereas the expansion of
\medskip

$x\in A\cap B$
\medskip

\noindent as
\medskip

$x\in A\ \wedge\ x\in B$
\medskip

\noindent is elementary.

\subsection{How the program works}

The following lines are taken directly from the program's code: they list, in order of priority, the names of the moves that it can do. In the rest of the section, we shall explain what those moves are.
\begin{verbatim}
  --Deletion
    deleteDone,
    deleteDoneDisjunct,
    deleteDangling, 
    deleteUnmatchable, 
  --Tidying 
    peelAndSplitUniversalConditionalTarget, 
    splitDisjunctiveHypothesis, 
    splitConjunctiveTarget, 
    splitDisjunctiveTarget,
    peelBareUniversalTarget, 
    removeTarget,
    collapseSubboxTarget,
  --Applying
    forwardsReasoning, 
    forwardsLibraryReasoning, 
    expandPreExistentialHypothesis, 
    elementaryExpansionOfHypothesis, 
    backwardsReasoning, 
    backwardsLibraryReasoning, 
    elementaryExpansionOfTarget, 
    expandPreUniversalTarget, 
    solveBullets,
    automaticRewrite,
  --Suspension
    unlockExistentialUniversalConditionalTarget, 
    unlockExistentialTarget,
    expandPreExistentialTarget,
    convertDiamondToBullet,
  --EqualitySubstitution  
    rewriteVariableVariableEquality,
    rewriteVariableTermEquality
\end{verbatim}

The program repeatedly applies a move of the first type it can from this list. Thus, if a move of type deleteDone can be performed, it performs it. If not, but a move of type deleteDoneDisjunct can be performed, then it performs that. Otherwise, it tries deleteDangling. And so on.


\subsubsection{Deletion moves}

The moves can be divided into some broad categories. We begin by discussing moves that delete statements from problem states.

Deletion is intended to model the human capacity to recognise that a statement is no longer going to be used and to turn attention away from it. Exactly how mathematicians do this involves several subtleties, which we shall discuss elsewhere. Here we give a brief indication of how the program works. One key point is that we do not normally delete statements unless they have been used as part of a reasoning step. This is because for the problems our program tackles, there is a strong presumption that every statement that appears will at some point be used, so it is risky to delete an unused statement, even if it appears to be unusable. However, once a statement has been used, this risk is greatly diminished, so under suitable conditions we are much happier to delete it. Therefore, the program tags hypotheses as `vulnerable' when they have been used, and its deletion rules are not applied to hypotheses unless they have this tag.

\subsubsection*{deleteDone}

There are several situations where a move results in a target being replaced by the word `done' because it has been proved. Once this has happened, the program immediately deletes it. The aim of the program is to reach a problem state with no targets.

\subsubsection*{deleteDoneDisjunct}

If a target is disjunctive and one of its disjuncts is the word `done', then the entire target is deleted.

\subsubsection*{deleteDangling} 

We call a free variable $v$ \emph{dangling} if it is involved in just one statement. This is a strong sign that the statement cannot be used. It is not a conclusive proof: for example, there might be an existential target that can be solved if we substitute $v$. However, if a statement contains a dangling variable and is vulnerable, then the program deletes it. This is a pragmatic decision on our part rather than one that is fully justified theoretically: there may be problems where deleting statements that have been used and that contain dangling variables is the wrong thing to do, but that situation does not appear to arise for the kinds of problems the program is designed to handle.

\subsubsection*{deleteUnmatchable}

Suppose that we have the statements $x\in A$ and $A\subset B$ as hypotheses. The expansion of $A\subset B$ is $\forall u\ u\in A\implies u\in B$. If we substitute $x$ for $u$, then the premise of this statement becomes $x\in A$, which is identical to the hypothesis. We say that $x\in A$ \emph{matches} the premise of (the expansion of) $A\subset B$. We call a statement \emph{unmatchable} if there are no available matches for it. 

The program is not allowed to substitute the same variable twice into the same hypothesis. (This is partly because no human would ever do so, and partly to avoid getting into a loop.) This can create further circumstances where a hypothesis is unmatchable. For example, suppose we apply forwards reasoning to the statements $x\in A$ and $A\subset B$ to deduce that $x\in B$. Then we can no longer use the match between $x\in A$ and $A\subset B$, so $x\in A$ becomes unmatchable (assuming that there is no other statement that matches it). Since it has been used, it is vulnerable, and will therefore be deleted. If no other statement matches $A\subset B$, then that too will be deleted.
\bigskip

\subsubsection{Tidying moves}

Tidying moves are moves that do not substantially change the logic of the problem state, but put it into a more convenient form.

\subsubsection*{peelAndSplitUniversalConditionalTarget}

If the target takes the form $\forall x\ P(x)\Rightarrow Q(x)$, then this move creates a new hypothesis $P(x)$ and replaces the target by $Q(x)$. This corresponds to the human move of saying (or thinking), `Let $x$ be such that $P(x)$; we need to show that $Q(x)$.' We can regard it as a composition of two moves: one to get rid of the quantifier and one to split up the conditional statement. We use the word `peel' to refer to any move that gets rid of a quantifier.
\bigskip

If there is more than one target, then this move has to be modified, since we cannot use $P(x)$ to help us prove a different target. In that situation, we create what we call a \emph{box}. A box looks like a problem state within a problem state: it has a line with some statements above it and some statements (or subboxes -- box formation can be iterated) below it. If the peel-and-split move creates a box, then that box lives below the main line.

That is, if we have a problem state of the form
\bigskip

Hypotheses

\longline

$\forall x\ P(x)\implies Q(x)$

$R$
\bigskip

then after the move we will transform it to
\bigskip

Hypotheses

\longline
\medskip

\begin{tabular}{|l|} \hline
$P(x)$ \\

\shorterline  \\

$Q(x)$  \\  \hline
\end{tabular}  
\medskip

$R$ 
\bigskip

The program then knows that it can use $P(x)$ to prove $Q(x)$ but not to prove $R$. The hypotheses above the main line can of course be used to prove both statements.

Boxes are regarded as complex targets (in the above case the target corresponding to the box would be to demonstrate that $P(x)$ implies $Q(x)$) and as such can form part of a list of targets. For example, if there are two boxes, one written above the other, that means that the implications within both boxes need to be established. Sometimes one needs to establish just one such implication: in that case one writes the two boxes side by side with a `$\vee$' symbol in between them.

In theory, boxes can be nested, though this rarely happens in practice. The rules governing which statements can be used to prove what are as follows. For each statement, there is a minimal box that contains it (counting the entire problem state as a box). Two boxes are \emph{comparable} if one contains the other in an obvious sense, and two statements are \emph{comparable} if the minimal boxes containing them are comparable. All deductions that the program makes must involve comparable statements.

A quick way to interpret a box is to regard the line in the middle as an implication sign, the lists of statements involved as linked by conjunction, and the box itself as placing brackets round everything.

\subsubsection*{splitDisjunctiveHypothesis}

If there is a hypothesis of the form $P\vee Q$, then the program splits the problem into two, one with $P$ as a hypothesis and one with $Q$ as a hypothesis. Technically, it achieves this by forming two boxes, one displayed above the other to indicate that both must be established. 

\subsubsection*{splitConjunctiveTarget}

If there is a target of the form $P\wedge Q$, then it is replaced by two targets $P$ and $Q$. 

\subsubsection*{splitDisjunctiveTarget}

If there is a target of the form $P\vee Q$, then it is replaced by two boxes, linked by a $\vee$ symbol. One box has nothing above the line and $P$ below the line, and the other has nothing above the line and $Q$ below the line. 

This move exists for technical reasons: for example, the program sometimes likes to attach tags to statements, but it has no facility for attaching tags to parts of statements. Therefore, if we want to use a tag to record information about one disjunct of a disjunctive target, we need to `split' the target first.

\subsubsection*{peelBareUniversalTarget}

If the target is of the form $\forall x\ P(x)$ and $P$ is not a conditional statement, then this move replaces the target by $P(x)$. 

There is a sense in which bare universal targets should never occur, since when we quantify over $x$, we do not quantify over every object in the universe, but rather over some set $X$. So one might argue that the statement $\forall x\ P(x)$ should really be rendered as the universal conditional statement $\forall x\ x\in X\implies P(x)$. What we mean by a `bare universal' statement is one where the `premises' are background statements that we do not want to elevate to `substantive' status. For example, we would regard the expansion of the statement `$G$ is Abelian' as a bare universal, since the only condition needed for two elements to commute is the background information that they are elements of $G$.

\subsubsection*{removeTarget}

This is actually a class of move types, but what they have in common is that under appropriate circumstances they replace a target with the word `done'. The most obvious example is when a target equals a hypothesis (and that hypothesis is allowed to be used to prove the target). A more complicated example is when the target is of the form $\exists u\ P(u)\wedge Q(u)$ and there are hypotheses $P(x)$ and $Q(x)$. The other circumstances are similar. 

\subsubsection*{collapseBoxedTarget}

If a box B has nothing above its internal line, contains no suspended variables (see below for a definition) and is not joined to another box with the $\vee$ symbol, then the statements below its internal line are listed as targets without the box B. 
\bigskip

\subsubsection{Applying moves}

An applying move is a move where we apply a hypothesis, result or definition. 

\subsubsection*{forwardsReasoning}

This covers a number of closely related steps that the program can take. The most basic example is using hypotheses of the form $P(x)$ and $\forall u\ P(u)\Rightarrow Q(u)$ to obtain the hypothesis $Q(x)$. However, there are several natural variants and generalizations of this. An obvious one is that if there are hypotheses of the form $P_1(x),\dots,P_k(x)$ and $\forall u\ P_1(u)\wedge\dots\wedge P_k(u)\implies Q(u)$, then the program generates the hypothesis $Q(x)$. Deductions of this kind are often called \emph{forward chaining}.

A variant that is worth highlighting is illustrated by the following simple piece of reasoning: if we know that $x\in A$ and that $A\subset B$ then we can deduce that $x\in B$. Humans will make this deduction in one step rather than first expanding the statement $A\subset B$ as $\forall u\ u\in A\Rightarrow u\in B$. Our program does the same. In general, for each type of reasoning move that involves a universal conditional hypothesis, there is a variant that does essentially the same thing to a pre-universal conditional hypothesis.

\subsubsection*{forwardsLibraryReasoning}

This is reasoning that is `mathematical' rather than `purely logical'. We have seen an example of it already: deducing from the statements `$(a_n)$ is a sequence in $A$', `$A$ is closed' and `$a_n\to a$' that $a\in A$. The reason this deduction can be done in one step is that the library contains a general result that says that whenever a sequence in a closed set tends to a limit, then the limit belongs to the closed set as well. 

Logically speaking, forwards library reasoning is similar to ordinary forwards reasoning, but there are one or two aspects of it that give it a different flavour. The main one is that library results contain no free variables: they are general facts that apply universally. This distinguishes them from hypotheses, which are more contingent. A second difference is that forwards library reasoning is normally used to deduce an atomic hypothesis from other atomic hypotheses. A universal conditional statement is involved, but it is in the library and is not a hypothesis. 

\subsubsection*{expandPreExistentialHypothesis}

As its name suggests, this means replacing a pre-existential hypothesis by its expansion. What the name of the move does not reveal is that this expansion is followed immediately by a peeling to get rid of the existential quantifier. So for example the statement `$\alpha$ has an inverse' might be replaced by $\alpha\beta=\beta\alpha=1$. 

This would be modelling a two-step human thought process. The first step is to note that there exists $\beta$ such that $\alpha\beta=\beta\alpha=1$, and the second step is quietly to forget about the existential quantifier and to refer to $\beta$ as though it has been chosen. Human mathematicians will usually miss out a step that says something like, `Let $\beta_0$ be such that $\alpha\beta_0=\beta_0\alpha=1$', so the program does as well.

\subsubsection*{elementaryExpansionOfHypothesis}

This takes a hypothesis that has an elementary expansion and replaces it by that expansion. This is sometimes combined with some tidying. For example, if the hypothesis in question is $x\in A\cap B$, then the elementary expansion is $x\in A\wedge x\in B$, but this expansion is immediately converted into the two hypotheses $x\in A$ and $x\in B$ and does not itself appear in any problem state.

\subsubsection*{backwardsReasoning}

Given a target $Q(x)$ and a hypothesis $\forall u\ P(u)\Rightarrow Q(u)$, this replaces the target by $P(x)$. 

More generally, if we have a target $Q(x)$ and a hypothesis $\forall u\ P_1(u)\wedge\dots\wedge P_k(u)\implies Q(u)$, then it is logically sound to replace the target $Q(x)$ by the $k$ targets $P_1(x),\dots,P_k(x)$. Deductions of this kind are often called \emph{backward chaining}. The program is allowed to do this more complex backward chaining only under tightly constrained circumstances: it must be that all but one of the statements $P_1(x),\dots,P_k(x)$ is a hypothesis, so that only one new target is created. This is another pragmatic decision: it is a crude way of deciding whether applying the hypothesis $\forall u\ P_1(u)\wedge\dots\wedge P_k(u)\implies Q(u)$ is likely to be the right thing to do, and the severity of the test is intended to stop the program making `speculative' deductions that risk leading to combinatorial explosion.

As with forwards reasoning, there is a simple variant where the role of the universal conditional hypothesis is played by a pre-universal conditional hypothesis instead. For example, given a target $x\in B$ and a hypothesis $A\subset B$ the program could use this variant to replace the target by $x\in A$.

\subsubsection*{backwardsLibraryReasoning}

This is backwards reasoning that makes use of a general result in the library. However, it is slightly subtler than forwards library reasoning, because it always uses hypotheses as well as a target. The precise rule is that if there are hypotheses $P_1(x),\dots,P_{k-1}(x)$, a library result $\forall u\ P_1(u)\wedge\dots\wedge P_k(u)=Q(u)$ and a target $Q(x)$, then the target can be replaced by $P_k(x)$. (The premises of the library result do not have to be stated in the order $P_1,\dots,P_k$.)

An example of this kind of reasoning would be to say, ``It is sufficient to prove that $B$ is open," if one wished to prove that $A\cap B$ was open and knew that $A$ was open. This would be making use of the result that an intersection of two open sets is open.

\subsubsection*{elementaryExpansionOfTarget}

This replaces a target by an elementary expansion of that target, if it has one. 

\subsubsection*{expandPreUniversalTarget}

This replaces a pre-universal target by its expansion. This move will be followed by one of the tidying moves peelAndSplitUniversalConditionalTarget or peelBareUniversalTarget. It is usually the first move that the program makes when faced with a naturally stated problem.

\subsubsection*{solveBullets}

As we are just about to discuss in more detail, we sometimes mark a variable $w$ with a diamond or a bullet. This indicates that the variable needs at some stage to be chosen in such a way that the problem can be solved. If the variable only ever appears in targets, then one simple way in which this can often be done is to identify another variable $x$ with the property that if we substitute $x$ for $w$, then every target that involves $w$ is equal to a hypothesis. In that situation, all those targets are replaced by `done'. This move is what we call `solveBullets'.

\subsubsection*{automaticRewrite}

There are a few rewriting rules stored in the library. Two examples are that the statement $x\in f^{-1}(A)$ is rewritten as the statement $f(x)\in A$ and the term $g\circ f(x)$ is rewritten as the term $g(f(x))$. (Of course, these are general rewriting rules and work whatever the variables happen to be called.) 

The rewriting of statements takes place only when those statements have been isolated as hypotheses or targets, so for example the program would not rewrite the statement $y\in f^{-1}(U)$ when it occurs inside the larger statement $\forall y\ d(x,y)<\delta\Rightarrow y\in f^{-1}(U)$. However, terms can be rewritten as soon as they appear. 
\bigskip

\subsubsection{Suspending moves}

We now come to the class of moves just alluded to: moves that help us deal with existential targets when it is not immediately clear what to substitute for the existentially quantified variable. A standard technique for this, which is essentially the technique we use, is to form \emph{metavariables}. The rough idea of a metavariable is that one reasons with it as though it had been chosen, deferring the actual choice until later when it becomes clearer what choice will make the argument work. Mathematicians often use this trick: a classic example is the `$3\epsilon$-argument' used to prove that a uniform limit of continuous functions is continuous.

When the program `pretends that it has chosen' a variable, it marks that variable with a bullet, and we say that it has been \emph{suspended}. Thus, suspension is the process of converting a variable into a metavariable. However, we found it convenient to introduce two `levels of suspension', to model two styles of reasoning that are logically similar but psychologically quite different. 

\subsubsection*{unlockExistentialUniversalConditionalTarget}

To illustrate this, suppose we have a target such as $\exists\delta\ \forall y\ d(x,y)<\delta\Rightarrow f(y)\in B$, and also a hypothesis $\forall u\ u\in A\implies f(u)\in B$. Then it is easy to see that we can reduce the target to $\exists\delta\ \forall y\ d(x,y)<\delta\implies y\in A$. However, this move is not open to the program because it is not allowed to `reason inside quantifiers'. This is a matter of convenience: such moves are logically valid, but it is tedious to specify appropriate variants of several of the reasoning moves listed above. Instead, we apply a procedure that we call \emph{unlocking}, which effectively moves aside the existential quantifier and allows the program to reason as normal with the statements inside it.

More precisely, what the program does to `unlock' the statement is create a box. In the example above, it would have no statements above the line, and below the line it would have the statement $\forall y\ d(x^\blacklozenge,y)<\delta\Rightarrow f(y)\in B$. The diamond on the variable $x$ indicates that $x$ needs to be chosen.

It is important for the program not to interchange quantifiers accidentally. For this reason, we tag the box just created with the variable $x^\blacklozenge$, to indicate that the existential quantification over $x$ is within that box. 

After unlocking the statement, the program peels and splits the resulting universal conditional target (so a more accurate name for the move type would be unlockPeelAndSplitExistentialUniversalConditionalTarget). After that, we have a box that looks like this.
\medskip

\begin{tabular}{|l|} \hline
$d(x^\blacklozenge,y)<\delta$ \\

\shortline \\

$f(y)\in B$ \\
\hline
\end{tabular}
\medskip

\noindent Once we have done this, the statement $f(y)\in B$ has become a target and the program is free to apply backwards reasoning to it.

\subsubsection*{unlockExistentialTarget}

This move replaces a target of the form $\exists x\ P(x)$ with a box that has nothing above the line and the statement $P(x^\blacklozenge)$ below the line. The box is labelled with the variable $x^\blacklozenge$. 

This move will never be applied to an existential universal conditional target, since that will have been dealt with by unlockExistentialUniversalConditionalTarget. The main reason we have two separate moves here is that we prefer to bundle the unlocking together with peeling and splitting when that is possible.

To see what unlockExistentialTarget allows the program to do, suppose that we have a target of the form $\exists x\ Q(x)\ \wedge\ R(x)$ and also a hypothesis of the form $\forall u\ P(u)\Rightarrow Q(u)$. In this situation we would like to be able to do backwards reasoning inside the existential quantifier to reduce the target to $\exists x\ P(x)\ \wedge\ R(x)$. However, the program does not have a move for this. Instead, it unlocks the existential target, so that it has a box with the statement $Q(x^\blacklozenge)\ \wedge\ R(x^\blacklozenge)$ below the line. The tidying move splitConjunctiveTarget can now turn this new target into two targets, and once it has done that, the applying move backwardsReasoning can be used to replace the target $Q(x^\blacklozenge)$ by $P(x^\blacklozenge)$. 

As another example of the use of unlocking, suppose that we wished to prove that $A\cap B$ is non-empty and had the hypotheses $x\in A$ and $x\in B$. The program cannot see that $x$ is a witness to the non-emptiness of $A\cap B$ without doing some processing. An obvious first step is to expand the target into the statement $\exists u\ u\in A\cap B$. However, the program is not then allowed to do an elementary expansion inside the quantifier. Instead, it unlocks $u$ so that there is a new target $u^\blacklozenge\in A\cap B$. This can now be expanded and split into the two targets $u^\blacklozenge\in A$ and $u^\blacklozenge\in B$, which solveBullets can then match with the hypotheses. 

This may seem a little circuitous, but it actually models quite closely how humans think. A human might say, `I want to show that $A\cap B$ is non-empty, so I need to find some $u$ that belongs to $A\cap B$. In other words, I need $u$ to be in $A$ and in $B$. Aha, I can take $x$.' The program's unlocking models the silent disappearance of the existential quantifier before the second sentence of the above.
\bigskip

\subsubsection*{expandPreExistentialTarget}

This does exactly what it says: it replaces a pre-existential target by its expansion.

\subsubsection*{convertDiamondToBullet}

There are certain moves that the program will not do with a `diamonded' variable. In particular, it will not do any reasoning with a hypothesis that involves such a variable: for that it needs a deeper level of suspension, roughly speaking corresponding to the human move of `pretending that a variable has been chosen' and then reasoning with it. Logically this is not an important difference, but it is a useful one for us because it reflects a difference in the way human mathematicians think and write. This helps the program to produce more convincing write-ups. 

We do not need separate move types for reasoning that involves hypotheses with bulleted variables: we just allow the reasoning moves above to take such variables. 

An important technicality is that if we postpone the choice of a variable, we must keep track of what other variables it is allowed to depend on. However, what we actually do is note which variables it is \emph{not} allowed to depend on. This is for two reasons. First, it seems to reflect more accurately how human mathematicians think about such variables, and secondly, it is more economical: there are typically many fewer variables on which a bulleted variable is not allowed to depend than variables on which it is allowed to depend.  

\subsubsection{Equality substitution}

If we are told that two objects are equal, then we can eliminate all mention of one object in favour of the other. The precise rules governing when and how mathematicians tend to avail themselves of this opportunity are not obvious. The rules below are best regarded as a temporary solution: they do not always result in realistically human choices, and we intend to replace them by more satisfactory rules in the near future.

\subsubsection*{rewriteVariableVariableEquality}

If there is a hypothesis of the form $x=y$, then this move replaces all occurrences of $y$ by $x$ and eliminates the hypothesis.

\subsubsection*{rewriteVariableTermEquality}

If there is a hypothesis of the form $v=t$ or $t=v$, where $v$ is a variable and $t$ is a term, then this move replaces all occurrences of $t$ by $v$. 

\bigskip

\subsection{Justification for the order of priority}

From examining how humans solve simple problems in the theory of metric spaces, it is not too hard to arrive at the above list of move types. But it is less clear what principles should govern the architecture -- that is, the way that the program decides which move type to do in any given situation. One obvious method of choosing an architecture is to work through a large number of problems and try to observe what seems to be the natural approach. After a while, one can make a guess at how the program should work, and if the guess results in strange behaviour for some problems then one can refine it, hoping that the process of refinement will stabilize quickly. An alternative method is to try to devise a theory that explains which move is the best than in each situation: ideally, that will turn out to be the move that humans are naturally drawn to. The second method is harder, but if it works, then the advantage is that the rules are likely to be more robust: without a proper theoretical backing, one cannot be as confident that they will not lead to inappropriate behaviour when the program is presented with an unfamiliar problem.

We have used a mixture of the two methods. We have good reasons for some of the choices we have made, but other choices are justified by the fact that they seem to work (in the sense of leading to human-like behaviour). A broad overarching principle that gives a theoretical backing to many of our choices is this: \emph{the program prefers safe moves to dangerous moves}. The picture we have here is one where at any stage there is a choice of moves that can be made, and we have to make an assessment of how likely any given choice is to form part of the argument one is looking for. The greater this likelihood, the safer the move.

Because it seems hard to attach probabilities to statements in this way, we have not tried to do so. However, since something like the above picture seems to be what humans do, we bear the picture in mind when planning the program. In particular, if a move is obviously safe, we will assign it a high priority.

A good example of a safe move is a tidying move. If, for example, we have a conjunctive hypothesis, then there is nothing to lose by splitting it up into its conjuncts, so that move we do automatically without any hesitation at all. By contrast, expanding a definition is substantially less safe: sometimes it is possible to reason in a high-level way without expanding, and since we do not allow `de-expansion' in this program (and in general allowing it would be highly problematic because of the danger of an infinite loop), expanding a definition is closing off the option of such high-level arguments. As an example, in the problem we discussed earlier, the program does not expand the statements `$(a_n)$ is Cauchy', `$(a_n)$ is a sequence in $A$' or `$(a_n)\to a$'. That allows it to do some high-level forwards library reasoning that would no longer be possible if any one of those three statements was expanded.

Thus, expansion has a fairly low priority. Having said that, some expansions, such as elementary expansions or expansions of pre-existential hypotheses, are considerably safer, so those ones have higher priority.

Somewhere in between are the other reasoning moves. Here it becomes more complicated to apply the general principle (even as an informal guiding principle), because the safety of a move type depends heavily on context. In particular, forwards reasoning is in general fairly unsafe -- if you have a lot of information and do not know which statements are relevant, then the probability that any given deduction will form part of the eventual proof may be quite small -- but is much safer when it comes to routine problems, which tend not to suffer from the problem of irrelevant information.

The psychology literature suggests that \emph{when it is safe}, humans tend to prefer forwards reasoning to backwards reasoning \cite{swelleretal1983,owensweller1985}, though this appears to be a question more of style than of problem-solving efficacy: we seem to prefer not to keep track of a moving target if we do not have to. Since forwards reasoning tends to be safe for the highly routine problems our program tackles, we have given all forwards reasoning a higher priority than all backwards reasoning. This also has the beneficial effect of making the program reluctant to switch direction -- too much switching from forwards to backwards or vice versa would again be bad mathematical style.

This aspect of our program is, however, unstable, in that we know that in order to develop the program we will have to change it. In fact, we even have an example of a rather routine problem where our program performs badly for this very reason. If it is asked to show that the intersection of two subgroups $H$ and $K$ of a group $G$ is itself a subgroup of $G$, then when it is trying to prove that $H\cap K$ is closed under multiplication, it expands, peels and splits the target, obtaining two elements $x$ and $y$ of $H\cap K$ and a target of showing that $xy\in H\cap K$. At that point, there is nothing to stop the program making `silly' deductions such as that $x^{-1}\in H$. There are various easy ways of dealing with this problem and we shall implement these in future versions: for example, we could add restrictions on creating new terms (a human would not think of deducing that $x^{-1}\in H$ because `we are not interested in $x^{-1}$') or we could alter the priorities so that when there are a number of possible forwards moves, so that it is no longer clear that they are all relevant, the program switches to backwards reasoning.

One other feature of the ordering of reasoning moves is that we prefer pure reasoning moves to library reasoning moves. That is because in general a hypothesis is more likely to be relevant than a library statement, though if enough of the premises of a library statement are present as hypotheses, that is a fairly strong argument for its relevance.

At the bottom of the list of priorities is suspension. That is because humans tend to regard it as a last resort. When mathematicians need to prove statements of the form $\exists x\ P(x)$, then by and large they prefer to simplify the problem until a suitable candidate $x_0$ for $x$ becomes obvious and it remains to carry out the relatively easy task of verifying that $P(x_0)$. Only when this straightforward approach fails do we take the more drastic step of pretending that $x$ has been chosen.

We will not say much more here about how we chose the priority order, but we have two brief further points. First, although our reasons are not 100\% precise, we found that in practice they were adequate, in the sense that they suggested an order before we started, and we found that we did not have to modify the order when we tried further problems (though, as commented above, there are certain aspects of the architecture that will need to be changed in future versions). Secondly, when it comes to the finer detail of the ordering, there may not be that much to choose between different move types. However, conflicts rarely arise between different move types that are not distinguished by any of the above criteria, so in practice these finer details have little if any effect on what the program actually does.

\subsection{Example: An intersection of two open sets is open}

Now that we have discussed how the program works, let us look at another example, which illustrates most of the move types and shows how the order or priority works in practice. The problem to be solved is the following.

\begin{problem}
Let $A$ and $B$ be open subsets of a metric space $X$. Prove that $A\cap B$ is open.
\end{problem}

\noindent The initial problem state is as follows.
\bigskip

$A$ is open

$B$ is open

\longline

$A\cap B$ is open
\bigskip

\noindent No reasoning moves are possible, so we end up having to expand. The highest priority move we can do is expandPreUniversalTarget, which, after the tidying move peelAndSplitUniversalConditionalTarget, has the following effect.
\bigskip

$A$ is open

$B$ is open

$x\in A\cap B$

\longline

$\exists\delta\ \forall y\ d(x,y)<\delta\Rightarrow y\in A\cap B$ 
\bigskip

\noindent It may look a little strange that we do not specify that $\delta>0$. The reason for that is that we think of positivity as a `background condition' rather than as a `substantive statement'. One reason that matters is connected with safety: if a premise of a universal conditional hypothesis is satisfied by some variable, then that is good evidence that the forwards reasoning one can do as a result is relevant to the problem. Or rather, it is good evidence if the premise is a `substantive statement', but not if it is merely a background condition such as positivity. So the distinction is a useful one. We think of background conditions as similar to, but not the same as, type declarations. The program does, however, know that $\delta$ is a positive real number: that fact is stored as background information when the statement `$A\cap B$ is open' is expanded.

At this point, the reluctance of the program to suspend $\delta$ means that it does as much forwards reasoning as it possibly can. It begins with elementaryExpansionOfHypothesis, applied to the third hypothesis.
\bigskip

$A$ is open

$B$ is open

$x\in A$

$x\in B$

\longline

$\exists\delta\ \forall y\ d(x,y)<\delta\Rightarrow y\in A\cap B$ 
\bigskip

\noindent This allows it apply two forwardsReasoning twice. After the first application, the problem state is as follows.
\bigskip

$A$ is open

$B$ is open

$x\in A$

$x\in B$

$\forall u\ d(x,u)<\eta[x]\Rightarrow u\in A$

\longline

$\exists\delta\ \forall y\ d(x,y)<\delta\Rightarrow y\in A\cap B$ 
\bigskip

\noindent Note that the last hypothesis is in a sense generated by a combination of submoves: the first is forwardsReasoning (using the hypotheses $x\in A$ and `$A$ is open') and the second is a peeling (to get rid of $\exists \eta$ at the beginning of the statement). However, the latter is so automatic that it is not listed as one of our tidying moves: instead, it is considered as part of any other move that potentially generates an existential hypothesis. 

It is important to keep track of the fact that $\eta$ depends on $x$, which is what is signified by $\eta[x]$. 

After this, deleteUnmatchable causes the program to delete the statements $x\in A$ and `$A$ is open'. This is because both statements have been used, so they are vulnerable, and because it is no longer permissible to substitute $x$ into `$A$ is open'. The resulting problem state is as follows.
\bigskip

$B$ is open

$x\in B$

$\forall u\ d(x,u)<\eta[x]\Rightarrow u\in A$

\longline

$\exists\delta\ \forall y\ d(x,y)<\delta\Rightarrow y\in A\cap B$ 
\bigskip

\noindent It then runs through a similar process for $B$ (it does not yet have the capacity to recognise that the problem is symmetric in $A$ and $B$ and say, `Similarly ...'). After that process, it arrives at the following.
\bigskip

$\forall u\ d(x,u)<\eta[x]\Rightarrow u\in A$

$\forall v\ d(x,v)<\theta[x]\Rightarrow v\in B$

\longline

$\exists\delta\ \forall y\ d(x,y)<\delta\Rightarrow y\in A\cap B$ 
\bigskip

\noindent It has now reached the point where it must suspend $\delta$. In the first instance, it uses the move unlockExistentialUniversalConditionalTarget, which includes a peeling and splitting. The result is as follows.
\bigskip

$\forall u\ d(x,u)<\eta[x]\Rightarrow u\in A$

$\forall v\ d(x,v)<\theta[x]\Rightarrow v\in B$

\longline
\medskip

\begin{tabular}{|l|} \hline
$d(x,y)<\delta^\blacklozenge[\overline y] $  \\

\shortline  \\

$y\in A\cap B$  \\ \hline
\end{tabular}
\bigskip

\noindent The notation $\delta^\bullet[\overline y]$ signifies that $\delta$ is not allowed to depend on $y$. 

The highest priority move the program can do now is elementaryExpansionOfTarget, so it does that, and automatically splits the resulting conjunctive statement (rather than using the move splitConjunctiveTarget).
\bigskip

$\forall u\ d(x,u)<\eta[x]\Rightarrow u\in A$

$\forall v\ d(x,v)<\theta[x]\Rightarrow v\in B$

\longline
\medskip

\begin{tabular}{|l|} \hline
$d(x,y)<\delta^\blacklozenge[\overline y]$ \\

\shortline  \\

$y\in A$  \\

$y\in B$ \\  \hline
\end{tabular}  
\bigskip

This allows it to apply backwardsReasoning twice. After the two deductions it reaches the following state. (It does them separately, so we are jumping a step here.)
\bigskip

{$\forall u\ d(x,u)<\eta[x]\Rightarrow u\in A$

$\forall v\ d(x,v)<\theta[x]\Rightarrow v\in B$

\longline
\medskip

\begin{tabular}{|l|} \hline
$d(x,y)<\delta^\blacklozenge[\overline y]$ \\

\shortline  \\

$d(x,y)<\eta[x]$  \\

$d(x,y)<\theta[x]$ \\  \hline
\end{tabular}  
\bigskip

\noindent It then uses deleteUnmatchable to delete the two hypotheses it has just used.
\eject

\longline
\medskip

\begin{tabular}{|l|} \hline
$d(x,y)<\delta^\blacklozenge[\overline y]$ \\

\shortline  \\

$d(x,y)<\eta[x]$ \\

$d(x,y)<\theta[x]$ \\   \hline
\end{tabular}
\bigskip

\noindent At this point, there is not much that the program can do, because it is not allowed to reason with the diamonded variable $\delta^\blacklozenge$. So the highest-priority move it can do is convertDiamondToBullet. Also, since there are no hypotheses above the main line, it may as well remove that line and the box.
\bigskip

$d(x,y)<\delta^\bullet[\overline y]$ 

\shortline

$d(x,y)<\eta[x]$ 

$d(x,y)<\theta[x]$ 
\bigskip

\noindent Now it applies backwardsLibraryReasoning. The result in the library is that if $a<b$ and $b\leq c$, then $a<c$. Applying that with the hypothesis and the first target results in the following problem state.
\bigskip

$d(x,y)<\delta^\bullet[\overline y]$

\shortline

$\delta^\bullet[\overline y]\leq\eta[x]$

$d(x,y)<\theta[x]$ 
\bigskip

\noindent The deletion rules do \emph{not} allow the program to delete the hypothesis we have just used (and this is a good example of a situation where deletion would be a very bad idea). However, it is aware that it cannot use the hypothesis with the new target. (We shall not describe here the precise mechanism by which it gains this awareness.) The program then uses backwardsLibraryReasoning again and this time it does delete the hypothesis, on the grounds that the variable $x$ that appears in the hypothesis is dangling. After that, it has reached the following state.
\bigskip

\shortline

$\delta^\bullet[\overline y]\leq\eta[x]$

$\delta^\bullet[\overline y]\leq\theta[x]$
\bigskip

\noindent This is a `standard' existence problem, so the solution is in a library of standard solutions and the program declares the problem solved. It is here that the background information that $\delta$, $\eta$ and $\theta$ are positive is used, since the library result is that the minimum of two positive real numbers $a$ and $b$ is a positive real number that is less than or equal to both $a$ and $b$.

That concludes the discussion of this example. For more examples of the program's output, as well as the precise output for this problem (which we have very slightly simplified in one or two places to make the presentation clearer) see \url{http://people.ds.cam.ac.uk/mg262/robotone.pdf} .

\subsection{Writing up}

In general, natural language generation is a complex process. It involves multiple levels of planning, which draw on both domain knowledge and models of the intended audience, and also a phase when the actual text is generated, which draws on syntactic, morphological and lexical information. An overview of the process may be found in \cite{reiter2000building}.  Because of this complexity, building a fully fledged natural language generation system is a major task.  Furthermore, since mathematics contains not just English words but also a large array of distinctive symbols used in distinctive ways, it is not at all straightforward to use off-the-shelf systems.  

Fortunately, mathematical language has properties that make the task considerably simpler than it is for the English language in general. Foremost among these is the fact that mathematical proofs almost always have a particularly simple rhetorical structure.  To some degree this is because the domain of discourse includes only timeless facts, which itself rules out a large proportion of the rhetorical relations found in general text.  But the main reason is that there is a strong convention that further constrains the rhetorical structure of proofs.  A proof proceeds by the presentation of a sequence of assertions, each of which follows from the premises of the theorem being proved or from previous assertions.  This structure is not accidental; it is a direct reflection of the fact that mathematicians process proofs by reading and verifying one sentence at a time, and would not expect the justification of a fact presented in one sentence to be deferred to a later sentence. (We are talking here about proofs of the level of simplicity of the proofs discussed in this paper. For more complicated arguments, facts may sometimes be used before they have been proved, but in good mathematical writing this will be carefully flagged up to make it as easy as possible for the reader to check that the resulting argument is complete and not circular.) 

This convention gives us an easy way to produce write-ups of our proofs. An obvious strategy is to allow each move to generate some number of sentences (possibly zero), and to concatenate output from the different moves to produce the final text.  Note that this strategy is viable only because we are absolutely rigorous about requiring our moves to reflect steps in human reasoning; in effect, the strategy is mimicking a human who is carefully writing down down a proof while coming up with it, which is quite straightforward for an experienced mathematician. (Again, this becomes less true if the proofs are more difficult.) As we shall see below, this simple strategy produces surprisingly good results, though with a weakness that needs to be dealt with by a postprocessing phase, which turns out to be straightforward. 

Because we have a fixed list of move types, implementing the strategy only requires us to specify which sentences (if any) are produced for the moves of each type.  A very simple way to do this is to use \emph{template generation}: each move type is associated with a \emph{template}, or `piece of text with holes', and the holes are filled in with concrete information about the facts and objects used in the actual move performed. So, for example, forwards reasoning may be associated with a very simple template `since $<\textit{facts}>$, $<\textit{deduced fact}>$'.  Instantiating this template would produce text like 
\begin{quote}
Since $A$\textrm{ is open} and $x\in A$, there exists $\eta > 0$ such that $u\in A$\textrm{ whenever }$\textit{d}(x,u) < \eta$.
\end{quote}
Note that individual facts are expressed in idiomatic ways, rather being displayed in a way that directly reflects the underlying predicate calculus; thus we have `$A$\textrm{ is open}' and `$\eta > 0$' rather than `$\textit{open}(A)$' and `$\textit{greater\_than}(\eta, 0)$'.  The same is true of objects: we display `$f \circ g$' rather than $\textit{compose(f,g)}$, and so on.  Similarly quantification is expressed idiomatically using words like `whenever', where possible, rather than using more stilted phrases like `for all', which would more directly reflect the underlying predicate calculus. 
\medskip

\noindent An example of the text produced by this method is as follows:
\begin{quote}
Let $x$ be an element of $A\cap B$. Since $x\in A\cap B$, $x\in A$ and $x\in B$. Since $A$ is open and $x\in A$, there exists $\eta>0$ such that $u\in A$ whenever $d(x,u)<\eta$. Since $B$ is open and $x\in B$, there exists $\theta>0$ such that $v\in B$ whenever $d(x,v)<\theta$. We would like to find $\delta>0$ s. t. $y\in A\cap B$ whenever $d(x,y)<\delta$. But $y\in A\cap B$ if and only if $y\in A$ and $y\in B$. We know that $y\in A$ whenever $d(x,y)<\eta$. We know that $y\in B$ whenever $d(x,y)<\theta$.
Assume now that $d(x,y)<\delta$. Since $d(x,y)<\delta$, $d(x,y)<\eta$ if $\delta\leq\eta$. Since $d(x,y)<\delta$, $d(x,y)<\theta$ if $\delta\leq\theta$. We may therefore take $\delta=\min\{\eta,\theta\}$. We are done.
\end{quote}

The main problem with this text is that it suffers a lack of coherence, in the sense defined in \cite{knott1996thesis}: the sentences are individually acceptable, but they do not combine to form an idiomatic discourse.  The principal reason for this is that the text repeats information unnecessarily.  For example, in 
\begin{quote}
Since $x\in A\cap B$, \underline{$x\in A$} and $x\in B$. Since $A$\textrm{ is open} and \underline{$x\in A$}, there exists $\eta > 0$ such that $u\in A$\textrm{ whenever }$\textit{d}(x,u) < \eta$.
\end{quote}
the repetition of the underlined phrase is awkward.  Because it is introduced by the sentence immediately preceding the `since' clause, it is awkward to have it spelt out explicitly within that clause.  Similarly, consider:
\begin{quote}
Since \underline{$\textit{d}(x,y) < \delta$}, $\textit{d}(x,y) < \eta$ if $\delta\leq \eta$. Since \underline{$\textit{d}(x,y) < \delta$}, $\textit{d}(x,y) < \theta$ if $\delta\leq \theta$.
\end{quote}
Here, having two identical `since' clauses in consecutive sentences is again awkward: the repetition of material is unwieldy and unidiomatic. 

We are of the opinion that \cite{knott1996thesis} correctly diagnoses the underlying problem here.  Spelling out rhetorical relations, or aspects of rhetorical relations, that can easily be inferred from the context violates Grice's maxim of quantity \cite{grice1975logic}.  Often the solution is to substitute an appropriate and less explicit \emph{cue phrase}.  For example,  `since $A$\textrm{ is open} and \underline{$x\in A$}, ...' is better replaced by `therefore, since $A$\textrm{ is open}, ...'. The cue phrase `therefore' (which assume the relevant reason has just been given) is less explicit than the cue phrase `since' (which subordinates an explicitly stated reason), so it avoids spelling out information that is clear from the context.  In other cases repetition can be avoided by combining sentences; for example, the second example may be changed into
\begin{quote}
Since \underline{$\textit{d}(x,y) < \delta$}, $\textit{d}(x,y) < \eta$ if $\delta\leq\eta$ and $\textit{d}(x,y) < \theta$ if $\delta\leq\theta$.
\end{quote}

The initial `sentence by sentence' process described above is followed by a series of transformations that manipulate pairs of consecutive sentences in order to resolve the issues just mentioned.  (Needless to say, the transformations operate on a structural level rather than on the literal text.)  Applying this series of transformations to the example text above yields:

\begin{quote}
Let $x$ be an element of $A\cap B$. Then $x\in A$ and $x\in B$. Therefore, since $A$ is open, there exists $\eta>0$ such that $u\in A$ whenever $d(x,u)<\eta$ and since $B$ is open, there exists $\theta>0$ such that $v\in B$ whenever $d(x,v)<\theta$. We would like to find $\delta>0$ s. t. $y\in A\cap B$ whenever $d(x,y)<\delta$. But $y\in A\cap B$ if and only if $y\in A$ and $y\in B$. We know that $y\in A$ whenever $d(x,y)<\eta$ and that $y\in B$ whenever $d(x,y)<\theta$.
Assume now that $d(x,y)<\delta$. Then $d(x,y)<\eta$ if $\delta\leq\eta$ and $d(x,y)<\theta$ if $\delta\leq\theta$. We may therefore take $\delta=\min\{\eta,\theta\}$ and we are done.
\end{quote}

One particular point worth emphasising is that the write-up process is deterministic: it will always produce the same output text for any given proof.  This is for two reasons.  First, if any non-determinism had been present we would have had to evaluate many outputs for any given proof, which would have made iterative improvement and fine-tuning of the write-ups considerably slower.  Secondly, and more importantly, if the process were nondeterministic, our claim that the program produced human-like output would be suspect, in that we would have been able to run the program several times and `cherry pick' output.  Unfortunately, this determinism has an undesirable (but fully anticipated) side-effect.  When one compare several proofs produced by the program, the write-ups are much more similar than those a human would produce.  For example, most proofs produced by the program end with the phrase `we are done'.  In the long run, we will undoubtedly need to introduce nondeterministic stylistic variation, allowing the program to vary the text generated for a particular step in just the way human would, despite the difficulties that will cause.

Finally, it is worth noting that during the evaluation process described in the next section, we collated a wealth of data on how humans write up proofs.  We anticipate using this data in combination with carefully chosen natural language processing techniques to create substantially improved versions of the write-up procedure.

\section{Testing the write-ups}

Once the program had generated the write-ups for several problems, we wanted to test whether they could pass for write-ups written by a human mathematician. In this section we describe an informal experiment that we carried out for this purpose. 

We began by asking two mathematicians, one an undergraduate and one a PhD student, to write out proofs for five problems for which our program had generated proof write-ups. We did not tell either of them why we were making this unusual request, and we did not ask them to make their write-ups as good as possible. One of the problems was to show that the inverse image of an open set under a continuous function is open, and one of our volunteers decided to prove the converse, so that he could use the topological definition of continuity to prove another of the assertions -- that a composition of continuous functions is continuous. We had to ask him to rewrite the latter and give the epsilon-delta proof, since we wanted the differences between the write-ups to be a matter of style rather than substance. 

We had another problem of this kind, which was that both our volunteers made frequent use of open balls. For example, their expansion of `$A\cap B$ is open' was `for every $x\in A\cap B$ there exists $\delta>0$ such that $B_\delta(x)\subset A\cap B$.' This made some of their arguments neater than the ones produced by our program. We contemplated getting the program to redo the problems using open-balls expansions, but in the end decided that it would be `cheating' to make changes to its output in response to the human write-ups we had solicited, so we left things as they were.

The program's write-ups were not designed to be indistinguishable from human write-ups: we merely wanted them to be acceptable as human write-ups. Therefore, we left in certain features, such as ending every proof with the words, `we are done', that we could with a little trouble have changed. (See the brief discussion of non-determinism at the end of the previous section.) For this reason, we did not want to ask people to guess which write-ups were the by the program. Instead, we presented all fifteen write-ups -- two by humans and one by the program for each of the five problems -- on the second author's blog, and asked readers of the blog to comment on them in any way they liked. We also asked them to award points for clarity and style. The orders of the write-ups were chosen randomly and independently. (The precise mechanism was to decide on a one-to-one correspondence between the set $\{1,2,3,4,5,6\}$ to the set of permutations of the set $\{1,2,3\}$, then to find a website that produced random dice rolls.) So that answers would be as independent as possible, all comments and ratings were sent to the blog's moderation queue and published only after the experiment was finished and comments on the blog post were closed. 

The post can be found at \url{http://gowers.wordpress.com/2013/03/25/an-experiment-concerning-mathematical-writing/} , together with all the comments and ratings, but the real point of the experiment was to see whether anybody noticed that not all the write-ups were by humans. Nobody expressed the slightest suspicion of this kind. 

Having said that, we should also remark that many commenters were highly critical of the program's output. Three criticisms in particular stand out. First, as we expected, the fact that the program did not use open balls was unpopular: many people commented that this made the write-ups unwieldy. Secondly, several of the human write-ups stated the new target when the initial one had been expanded, peeled and split. Several readers commented that they found this helpful, and criticized our program for not doing it. And thirdly, commenters did not like the way the program spelt out in detail how it thought of the right variable to substitute into existential targets (such as choosing $\min\{\eta,\theta\}$ for $\delta$ in the intersection-of-open-sets problem. 

It would be easy to modify the program so that none of these criticisms apply, so they do not point to fundamentally non-human aspects of how it thinks. To change the first, we would just have to use open-balls expansions of definitions such as `$A$ is open' and `$f$ is continuous'. To change the second, we could alter the rule for what the write-up does when we expand, peel and split, so that it states the new target (preceded by a phrase such as `We need to show that'). The third criticism would be harder to deal with, but when we switch to having two styles of write up -- a `proof write-up' and a `proof-discovery account' -- then for the first style we will let the program work out the values of bulleted variables, then simply declare those values when the variable is first mentioned after being suspended. This will correspond to the human practice of writing something like `Let $\delta=\min\{\eta,\theta\}$' or `Consider the sequence $(b_n)$ defined by $b_n=a_n/(1+a_n)$,' which `magically' does exactly what it needs to do later in the proof. 

Although our program's output came in for quite a bit of criticism, so did the write-ups by the undergraduate and PhD student -- it seems that the readers were harsh judges. However, for most of the problems, the human write-ups were found preferable to the program's.

After the success (as we considered it) of this experiment, we dared to try a direct test. We published a new post, this time explaining that one proof was by a program, one by an undergraduate and one by a PhD student, and inviting readers to vote on which one they thought was by the program. For each problem, the write-ups were numbered (a), (b) and (c). There were seven options for the voting: one could choose between (a), (b) and (c), but also choose between `The computer-generated output is definitely $*$' and `I think the computer-generated output is $*$ but am not certain'; the seventh option was `I have no idea which write-up was computer generated.' Again there was the opportunity to comment, for those who wanted to explain the reasons for their choices.

We did not reveal the results of the voting so far, or anybody's comments, until the experiment was ended and the voting was closed. However, there was a different kind of dependence between answers, which was that people had the opportunity to look for clues that two different write-ups were from the same source. Given that we had not tried to remove stylistic `tics' from our program's write-ups, this put the program at a significant disadvantage. It was clear from the comments that many people had noticed that for each problem exactly one write-up ended with the words `we are done'. 

Despite this, the program did reasonably well at fooling people that it was human. The typical pattern was that roughly half the voters would correctly guess which output was by the program, with slightly under half of that half saying that the output was definitely by the program. The undergraduate would always `come second', and there would always be a fair number of people who said that they had no idea which output was written by the computer. There were surprisingly many votes for `The computer-generated output is definitely $*$,' when $*$ was the wrong answer. The total number of votes was always at least 300, and for the first problem listed (the intersection of open sets is open) it was over 1000. One slight complication was that after a day or two the post was listed on the front page of Hacker News. The result was that the number of votes doubled in a couple of hours, and it may be that the profile of a typical voter changed. Fortunately, we had noted down the voting numbers just before this happened, so we presented those results as well as the final numbers. In the end, however, the proportions did not change very much. The detailed numbers can be found here: \url{http://gowers.wordpress.com/2013/04/14/answers-results-of-polls-and-a-brief-description-of-the-program/} .

One thing this experiment could not tell us, except to a limited extent through the comments, was whether the program was good at fooling \emph{mathematicians} that it was human. It could be that the more mathematically experienced readers found the program's output easy to distinguish, while the votes for the human write-ups came from people who were not used to reading mathematical proofs. However, we feel justified in concluding that the program's output is not \emph{obviously} written by a computer program, and that was our objective. 

\section{Future work}

In the long term, we would like to enlarge significantly the set of problems that our program, or some new version of it, is capable of solving. To do this, we will have to enable the program to handle certain kinds of deductions that it currently handles either not at all or only in a rather rudimentary way. In particular, an immediate target is to give the program the means to deal with second-order quantification, which would allow it to solve easy compactness problems, and also problems that require the construction of `obvious' sequences.

At a more basic level, the program does not currently solve problems that involve proof by contraposition or contradiction. It is not hard to add moves that allow it to cope with a few problems of this kind, but it is trickier to do so while not letting it apply those moves in inappropriate contexts. More work is needed to understand what triggers the `contradiction move' in human mathematicians, but we expect to be able to add this facility in the near future. 

The program is also not as good as we would like at handling equality substitutions. The situation here is similar: we can obviously add moves that do such substitutions (and have done so in the current version of the program), but it is more challenging to understand when humans make such substitutions. It is also tricky to come up with a general understanding of how they choose which out of two equal variables or complex terms to eliminate. At its most general, the problem of how to handle equality is well known to be hard, but our immediate aim would be a program that can handle the easy cases of that problem competently and in a human way.

\bibliographystyle{spmpsci}
\bibliography{robot}

\end{document}